\documentclass{article}
\usepackage{spconf,amsmath,graphicx,hyperref}
\usepackage{booktabs}
\usepackage{multirow}
\usepackage{tabularx}

\usepackage{xcolor}

\title{See No Evil: Semantic Context-Aware Privacy Risk Detection for AR }

\name{Jialu Liu$^{\star}$ \qquad
  Yao Li$^{\star}$ \qquad
  Zhuoheng Li$^{\star}$ \qquad
  Huining Li$^{\dagger}$ \qquad
  Ying Chen$^{\star}$
}
\address{
  $^{\star}$Pennsylvania State University, State College, PA \\
  $^{\dagger}$North Carolina State University, Raleigh, NC
}

\begin{document}

\maketitle

\begingroup
\renewcommand\thefootnote{}
\footnotetext{\textcopyright\ 2026 IEEE. Personal use of this material is permitted. Permission from IEEE must be obtained for all other uses, in any current or future media, including reprinting/republishing this material for advertising or promotional purposes, creating new collective works, for resale or redistribution to servers or lists, or reuse of any copyrighted component of this work in other works.}
\addtocounter{footnote}{-1}
\endgroup

\begin{abstract}

Augmented reality (AR) systems pose unique privacy risks due to their continuous capture of visual data. Existing AR privacy frameworks lack semantic understanding of visual content, limiting their effectiveness in detecting context-dependent privacy risks. We propose PrivAR, which leverages vision language models (VLMs) with chain-of-thought prompting for contextual privacy risk detection in AR environments. PrivAR uses visual scene cues to infer potential sensitive information types, such as identifying password notes in office environments through contextual reasoning.
PrivAR detects and obfuscates textual content, preventing exposure of sensitive information while preserving contextual cues necessary for VLM inference. Additionally, we investigate contextually-informed warning interfaces to enhance user privacy awareness. Experiments on a real-world AR dataset show that PrivAR achieves superior accuracy (81.48\%) and F1-score (84.62\%) compared to baselines, while reducing privacy leakage rate to 17.58\%. User studies evaluating contextually-informed warning interfaces provide insights into effective privacy-aware AR design. 

\end{abstract}

\begin{keywords}
Augmented reality, privacy risk detection, vision language model, scene understanding

\end{keywords}

\section{Introduction}
\label{sec:intro}

Augmented reality (AR) 
augments users’
perception by overlaying digital information onto the physical world around the users. 
AR relies on ``always-on" environmental sensing, where cameras continuously capture users' surroundings~\cite{291287}. This poses unprecedented privacy challenges by inadvertently recording sensitive information about users and non-consenting bystanders~\cite{paneva2025user}. 

Approaches to addressing AR privacy threats evolve from basic access controls to sophisticated countermeasures.
``All-or-nothing" access controls from traditional mobile systems, such as camera toggles, are incompatible with AR's always-on sensing requirements~\cite{o2023privacy}.
Recent frameworks move toward context-aware, device-level control. In~\cite{roesner2014world}, world-driven access control is introduced as an environment-centric control model that enables real-world objects 
to broadcast access policies. 
Erebus~\cite{291287} introduces a domain-specific language for declaring the functional intent of an application and implementing granular permission controls. VACMaps~\cite{zhu2022verifiable} ensures provably correct access control
system for localization and mapping 
through a formal verification engine.
BystandAR~\cite{corbett2023bystandar} 
employs the device user’s eye gaze and voice
to detect 
bystanders, applying real-time 
visual distortion to protect bystanders' information. 
Despite these advances, current AR privacy frameworks focus mainly on 
 access control policies and categorical object identification. To complement this work, we focus on context-dependent privacy risk detection enabled by semantic understanding of visual content.

Prior research has focused on analyzing visual data 
to detect and mitigate privacy risks, employing computer vision techniques for specific object classes~\cite{Segue,kuang2024facial} as well as advanced vision language models (VLMs) for contextual understanding~\cite{murrugarra2025beyond,mishra2025revisiondatasetbaselinevlm,chen-etal-2025-vision}.
Computer vision techniques detect sensitive content (e.g., faces, license plates) and apply anonymization filters~\cite{
Segue, kuang2024facial}, though aggressive obfuscation degrades utility for downstream tasks~\cite{kuang2024facial}. 
VLM-based approaches 
have potential for semantic interpretation of visual content.
FiG-Priv~\cite{murrugarra2025beyond} employs 
VLMs for personally identifiable information 
redaction. ReVision~\cite{mishra2025revisiondatasetbaselinevlm} uses on-device VLMs to rewrite multimodal instructions as text-only commands. VisShield~\cite{chen-etal-2025-vision} guides VLM to perform optical character recognition (OCR) for healthcare text de-identification. 
However, VLM-based systems face the ``recursive privacy problem" where sensitive images must be shared with the VLM itself. Hence, we perform 
information obfuscation 
before sending data to the VLM.

Privacy-preserving AR also requires usable privacy warnings in AR interaction.
User studies reveal common misunderstandings in AR privacy, such as 
assuming location data is needed for distance measurement~\cite{paneva2025user}.  
As a result, users may make uninformed decisions about granting or rejecting permissions. 
Schaub et al.~\cite{schaub2015design} show that effective privacy warnings must be contextually relevant. 
This motivates our investigation of how different warning modes, informed by privacy risk detection, can improve privacy awareness in AR. 

In this paper, we propose PrivAR 
for contextual understanding of AR environments to enhance privacy risk detection. Leveraging VLMs, PrivAR uses visual scene cues (e.g., users are experiencing AR in offices) to infer potential sensitive information (e.g., password notes in offices). We address the recursive privacy problem 
in VLM-based systems
by private information obfuscation. We also investigate how contextually-informed warning interfaces can enhance user privacy awareness in AR environments.

Our contributions include: (1) We enhance privacy risk detection by 
understanding AR environments and inferring 
sensitive information types 
from visual scene cues, and subsequently obfuscate 
sensitive information. 
 (2) We design contextually-informed privacy warning interfaces. 
 (3) We conduct dataset-based experiments and user studies to evaluate PrivAR. Our code and dataset are publicly available.~\footnote{https://github.com/jlliu2001/AR-Privacy-Detection}

\begin{figure}
    \centering
    \includegraphics[width=0.95\linewidth]{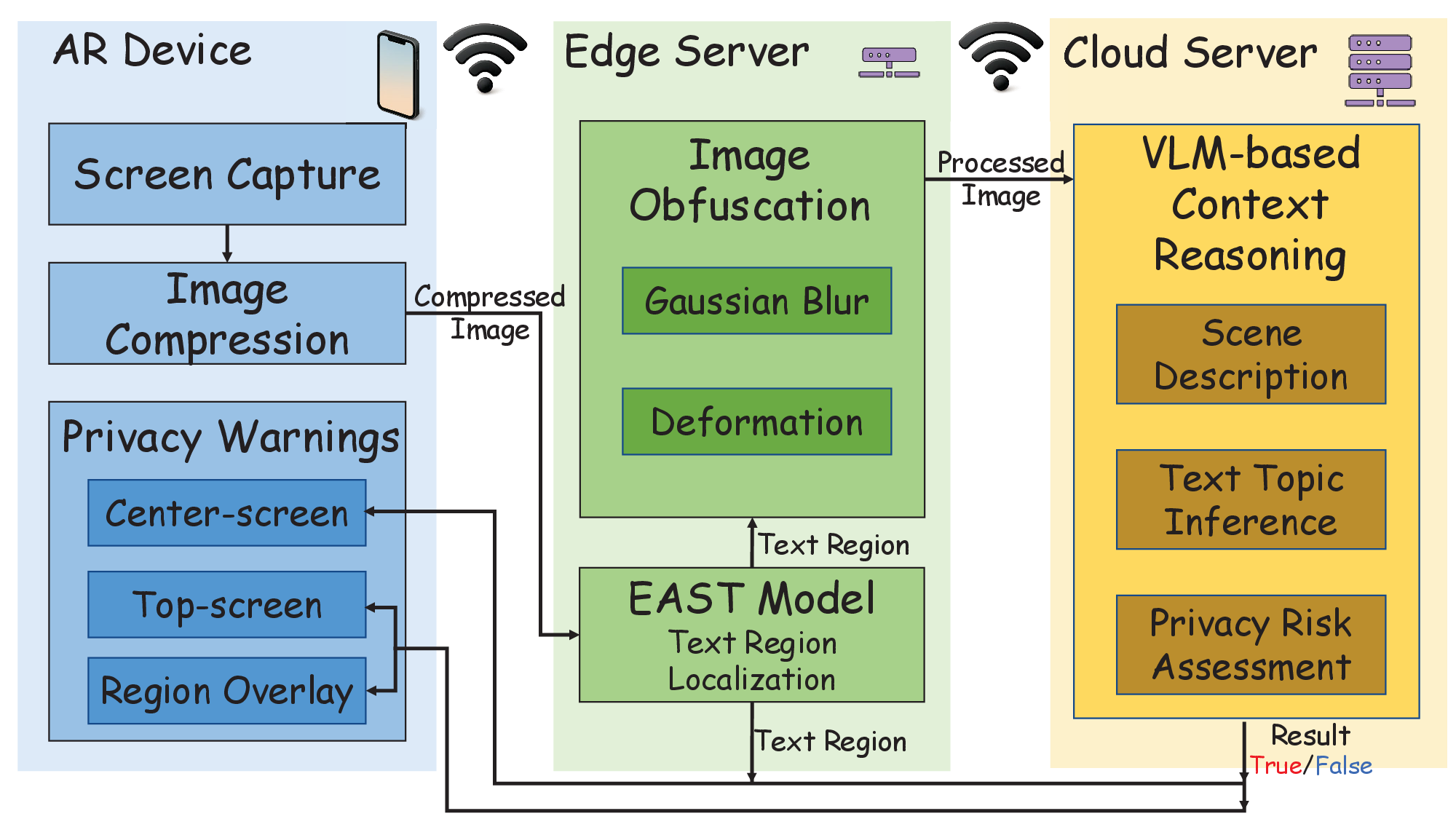}
    \caption{System architecture of PrivAR. 
    }
    \vspace{-0.5cm}
    \label{fig:fig1}
\end{figure}

\section{SYSTEM DESIGN}
\label{sec:system_design}

As shown in Fig. \ref{fig:fig1}, PrivAR has a 
three-tier architecture comprising: (1) an AR device providing user interface and privacy risk warnings (Section~\ref{sssec:device}), (2) an edge server handling private information obfuscation (Section~\ref{sssec:edge}), and (3) a cloud server inferring contextual information and performing privacy risk assessment (Section~\ref{sssec:cloud}). In PrivAR, nearby computing units (e.g., personal computers) serve as edge servers, consistent with prior studies~\cite{younis2020latency,scargill2022integrated} and commercial AR systems (e.g., Varjo XR-4, Meta Quest with Air Link). 
Data is transmitted wirelessly between three components.

\subsection{
User Interface for Privacy Risk Warning}
\label{sssec:device}

\begin{figure}
    \centering
    \includegraphics[width=0.95\linewidth]{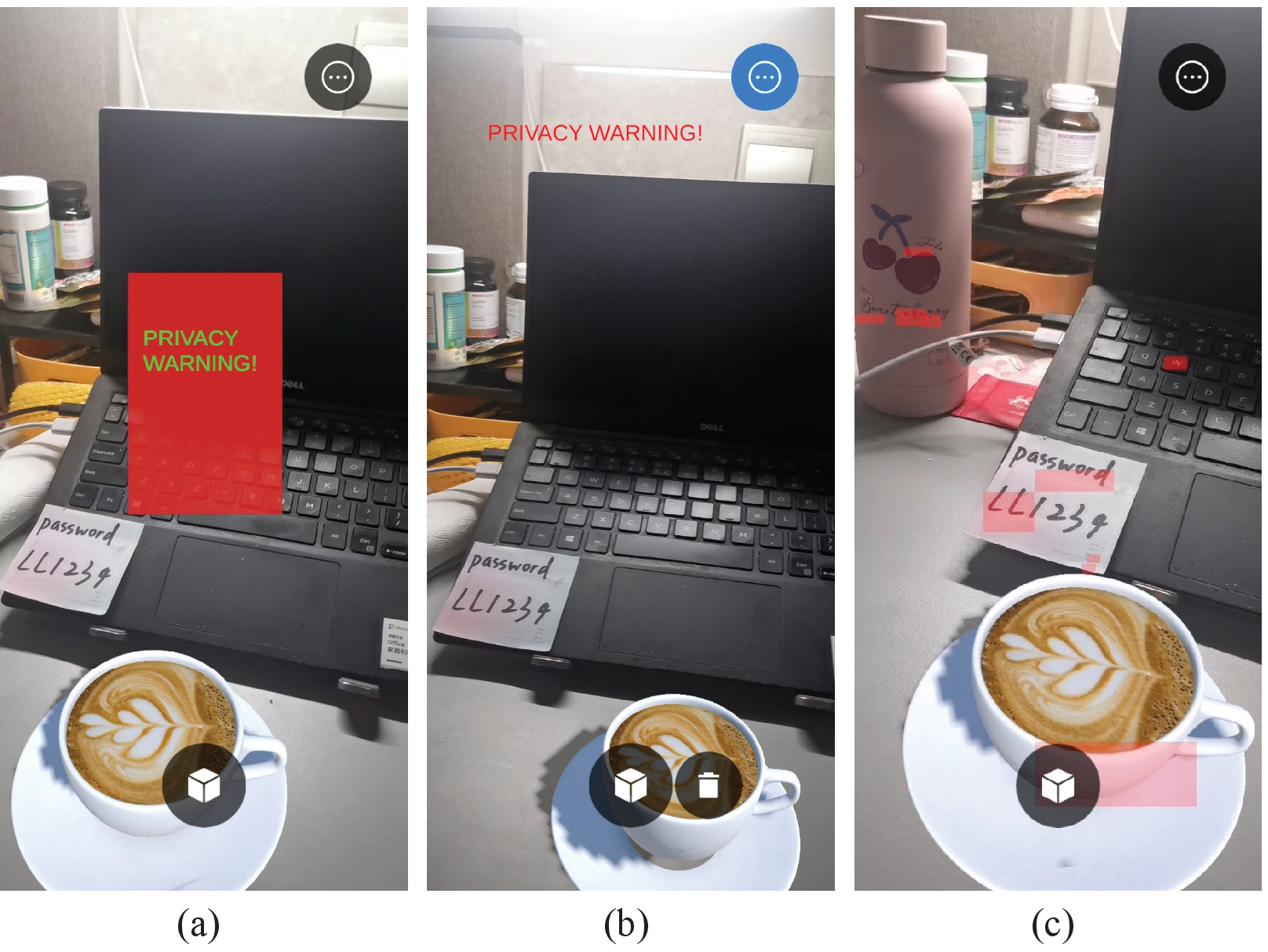}
    \caption{Different warning modes when a risk of privacy leakage is identified: (a) center-screen warning, (b) top-screen warning, and (c) region overlay warning.}
    \label{fig:fig2}
    \vspace{-0.4cm}
\end{figure}

The AR device continuously captures images of the surrounding environment, 
which are compressed to reduce transmission latency to the edge server and subsequently to the cloud server. The AR device then receives the privacy risk detection results from the cloud. 
When sensitive data is detected (Section~\ref{sssec:cloud}), users are alerted with privacy risk notifications to help safeguard their privacy during AR use. 
We design 3 warning modes, 
all 
flashing in a 2-second cycle (one second on, one second off) for a total of 6 seconds, as shown in Fig.~\ref{fig:fig2}. (1) With \emph{center-screen warning}, 
a red rectangle with “PRIVACY WARNING!” text appears at the center of the AR interface.
(2) \emph{Top-screen warning} shows red text stating “PRIVACY WARNING!” displayed at the top of the AR interface.
(3) With \emph{region overlay warning}, 
red rectangles are overlaid on the regions detected as containing private information.

\subsection{
Private Information Obfuscation} 
\label{sssec:edge}

The edge server receives compressed images $I_\mathrm{comp}$ from the AR device and applies a lightweight text detection model, denoted as $D_\mathrm{text}$, to localize text regions. We select EAST~\cite{zhou2017east} for text detection
due to its high efficiency and sufficient accuracy compared with more advanced methods. EAST outputs a set of $n$ bounding boxes, $ B= \{b_1, b_2, \ldots, b_n\} = D_\mathrm{text}(I_\mathrm{comp})$, where each $b_i$ defines a rectangular text region in $I_\mathrm{comp}$. 
A text obfuscation function $\mathcal{O}$ is then applied to the pixels within these bounding boxes. 
$\mathcal{O}$ combines Gaussian low-pass filtering and spatial elastic deformation. Gaussian blur is denoted as $G(I_\mathrm{comp}; \sigma)$, where $\sigma$ controls the blur intensity. Elastic deformation $E(G(I_\mathrm{comp}; \sigma);\beta)$ introduces spatial perturbations via a random warp field of scale $\beta$. This 
suppresses high-frequency textual details and introduces non-linear geometric distortions, 
while preserving overall object morphology. 
 Let $\mathcal{M}$ be a binary mask generated from $B$. 
 The final obfuscated image, $I_\mathrm{obf}$, is given by
$I_\mathrm{obf} =\mathcal{O}(I_\mathrm{comp};\sigma,\beta) \notag =(1-\mathcal{M})\odot I_\mathrm{comp}+\mathcal{M}\odot E(G(I_\mathrm{comp},\sigma);\beta))$,
where $\odot$ denotes element-wise multiplication.
Processed images are then sent from the edge server to the VLM in the cloud 
for performing various downstream tasks. 
In our design, these AR images are used by the cloud server for privacy risk detection. 

\subsection{Contextual Inference and Privacy Risk Assessment}

\label{sssec:cloud}

The cloud server hosts the VLM, leveraging its 
reasoning capabilities to analyze obfuscated images and deliver privacy risk assessments for display on the AR device. 

Since textual regions are intentionally obfuscated, 
we directs the VLM to use contextual cues such as scene features and local visual information near the obfuscated areas to infer the likely content of the obfuscated text.
We employ the chain-of-thought (CoT) prompting strategy~\cite{COT}
that guides the VLM through a three-stage reasoning pipeline.

\noindent\textbf{Scene description.} Given the obfuscated image, 
the VLM first analyzes visual features to classify the AR environmental context 
(\textit{``The user is experiencing AR in an office setting with a desk and a computer''}). This contextual understanding helps to narrow down the range of possible private information 
present in the AR environment.

\noindent\textbf{Text topic inference. }
Based on the semantics revealed by local pixels near the obfuscated text regions and the overall environmental context, the VLM conjectures the likely topic of the obfuscated text. 
For instance, if the scene is recognized as a kitchen (from the scene description step) and the obfuscated text is located on a sheet of paper attached to the refrigerator, the VLM may be prompted to infer: \emph{“Given that the scene is a kitchen and that the text appears on a paper on the fridge, the text is likely a grocery list or a recipe.”}

\noindent\textbf{Privacy risk assessment.} Finally, the VLM combines the inferred text topic with the environmental context to determine whether the content contains private information, generating a binary risk assessment. 
For example, if the scene is recognized as an office and the inferred text is a meeting agenda displayed on a whiteboard, the VLM may reason: \emph{“The scene is an office, and the text is a meeting agenda. Since meeting agendas can contain sensitive information, capturing this may constitute a privacy risk.”}

\section{EVALUATION}

We demonstrate the privacy risks of AR applications by evaluating the accuracy of privacy risk detection. We also assess the effectiveness of our privacy-preserving approach.

\begin{table*}[]
\centering
\caption{Privacy risk detection performance of PrivAR and baseline approaches.}
\label{part1_result}
\begin{tabular}{
  >{\centering\arraybackslash}p{50mm}
  >{\centering\arraybackslash}p{18mm}
  >{\centering\arraybackslash}p{18mm}
  >{\centering\arraybackslash}p{18mm}
  >{\centering\arraybackslash}p{18mm}
}
\toprule
{Detection method}    & {Acc.(\%)$\uparrow$} & {Prec.(\%)$\uparrow$} & {Rec.(\%)$\uparrow$} & {F1(\%)$\uparrow$} \\ \midrule
Rule-based                   & 39.58                 & 44.00                  & 8.63                & 14.43           \\
Sensitive object recognition & 55.79                 & 50.00                  & 83.77               & 62.62           \\
Scene captioning-based       & 67.36                 & 82.02                & 57.25               & 67.44           \\ 
PrivAR (ours)                         & \textbf{81.48}                 & \textbf{83.02}                  & \textbf{86.27}               & \textbf{84.62}           \\
\bottomrule
\end{tabular}
\vspace{-0.5cm}
\end{table*}

\subsection{Experimental Setup}
\label{ssec:subhead}

We conduct 
a dataset-based evaluation and a user study with our end-to-end AR system.
Following the experimental methodology in~\cite{xiu2025viddar}, in the dataset-based evaluation, 
we collect a dataset of 432 screenshots, where each screenshot corresponds to one real-world AR interaction using our end-to-end system in 4 physical scenes (office, living room, bedroom, and café).
The 
images cover 
6 private information types (ID cards, credit cards, password notes, transcripts, medical reports, and text displayed on computers or phones), and 6 virtual objects in AR interactions (coffee cup, whiteboard, indoor plant, guitar, vase, and chair). 40.63\% of positive samples contain multiple sensitive items (e.g., handwritten password notes and transcripts together). There are 94 images containing negative samples that are visually similar to sensitive content but are non-sensitive, 
including published papers. 
Our AR system was implemented in Unity 2022.3.6f1, using a Google Pixel 10 (Android 16) as the AR device.
A Dell 16 Premium laptop (16GB RAM, Intel Core Ultra 7 CPU) serves as the edge server.

By default, we use GPT-4o mini~\cite{achiam2023gpt} as the VLM and the text-only LLM. We also study the impacts of different VLMs (GPT-4o mini, GPT-4o, and LLaMA-4-Maverick-17B-128E-Instruct~\cite{LLaMA}). 
We empirically set the image compression ratio to 75\%, 
$\sigma = 5$, and 
$\beta = 40 $ in Section~\ref{sec:system_design}.

\noindent\textbf{Privacy risk detection performance.} 
To assess the effectiveness of the privacy risk detection module in PrivAR, we implement and compare several baseline methods representing major existing paradigms:
(1) \emph{Rule-based approach} employs EAST~\cite{zhou2017east} and OCR~\cite{smith2007overview} to identify text regions and extract textual content. The extracted text is then analyzed using 
predefined patterns to identify structured sensitive information such as ID numbers, phone numbers, and credit card numbers. 
(2) \emph{Sensitive object recognition approach} uses YOLOv8~\cite{yolov8} fine-tuned on our dataset to identify objects commonly associated with privacy risks, such as laptops and ID cards.
(3) \emph{Scene captioning-based approach} avoids using raw visual data, thereby reducing exposure of private information. 
Instead, this approach uses the VLM 
to generate a comprehensive textual scene description (i.e., caption) for each obfuscated image. This caption is then used by a text-only LLM to make the final privacy risk detection.
 
Privacy risk detection is a binary classification task, where each image is classified as either containing private information or not.
Detection performance is evaluated using accuracy, precision, recall, and F1-score on the collected dataset.

\begin{table}[t]
\centering
\caption{Privacy risk detection performance of PrivAR with different VLMs.}
\label{part2_result}
\vspace{0.1cm}
\begin{tabular}{
  >{\centering\arraybackslash}p{23mm}
  >{\centering\arraybackslash}p{10mm}
  >{\centering\arraybackslash}p{11mm}
  >{\centering\arraybackslash}p{10mm}
  >{\centering\arraybackslash}p{10mm}
}
\toprule
VLM & Acc.(\%)$\uparrow$ & Prec.(\%)$\uparrow$ & Rec.(\%)$\uparrow$ & F1(\%)$\uparrow$ \\
\midrule
GPT-4o mini      & 81.48                 & 83.02                  & 86.27               & 84.62 \\
GPT-4o           & \textbf{82.87}          & \textbf{84.41}          & \textbf{87.06} & \textbf{85.71} \\
LLaMA-4-Mav. 
& {78.47}
& {81.15}
& {82.75}
& {81.94} \\
\bottomrule
\end{tabular}
\vspace{-0.5cm}
\end{table}

\label{sssec:subsubhead}

\noindent\textbf{Privacy preservation effectiveness.}
To evaluate the effectiveness of our privacy preservation methods, we examine how much sensitive information can be extracted from AR images using both OCR and VLM evaluation.
For OCR evaluation, we compare text extracted from original and 
obfuscated images using the character error rate (CER) metric~\cite{ColOLHTR}. 
For VLM-based evaluation, we present both original and obfuscated images to the VLM, prompting it to identify any sensitive content. Privacy leakage is determined when the VLM detects the same sensitive information in both 
images. The privacy leakage rate (PLR) is calculated as the proportion of image pairs where this exact match occurs after obfuscation.

We examine two alternative methods. (1)
 \emph{PriAR w/o obfuscation:} The captured image is compressed and sent 
 to the VLM for privacy risk detection, without applying any obfuscation. This method reflects the performance of PriAR when privacy-preserving processing is disabled. 
(2) \emph{Oracle-guided obfuscation:} Text regions are blurred according to ground-truth annotations rather than 
text region detection, ensuring precise removal of sensitive information. The VLM then performs privacy risk assessment on these ideally obfuscated images. This approach serves as an upper bound for privacy protection, representing the best-case scenario for our method.

\noindent\textbf{User Study.} 
We conduct an institutional review board (IRB)-approved user study with 10 participants 
to assess AR privacy risk awareness, and preferences for warning modes (shown in Fig.~\ref{fig:fig2}) in AR interfaces. 
Participants 
completed 
questionnaires 
about concern over privacy leakage from AR applications (pre- and post-experiment), trust in our PrivAR system for alerting users to privacy risks (post-experiment), and preferred warning method (post-experiment).

\subsection{Evaluation Results}
\label{sec:typestyle}

\textbf{Privacy risk detection performance.}
Table~\ref{part1_result} summarizes the performance of PrivAR and three baseline methods on privacy risk detection. PrivAR 
achieves the best overall performance, with 81.48\% accuracy, 83.02\% precision, 86.27\% recall, and 84.62\% F1-score. 
The rule-based approach shows 
low recall (8.63\%) and accuracy (39.58\%). This indicates that a rule-based approach frequently misses actual privacy risks, especially when text formats deviate from its predefined patterns and rules.
Sensitive object recognition has a 55.79\% accuracy, likely due to the presence of hard negative samples, where visually similar non-sensitive objects are frequently misclassified as sensitive. Scene captioning-based method achieves relatively high precision, yet exhibits lower overall accuracy than PrivAR. 
The low recall of the scene captioning-based method 
indicates frequent false negatives, as captions typically capture only salient scene elements while omitting subtle, partially visible, or context-dependent privacy cues. 
Overall, these results indicate that 
PrivAR can reliably identify diverse 
privacy risks (including those that are subtle or context-dependent) across different AR scenes.
 
\noindent\textbf{Impact of different VLMs used in PrivAR.}
Table \ref{part2_result} shows that PrivAR consistently achieves high privacy risk detection performance when integrated with 
VLMs of varying sizes and architectures. While GPT-4o attains slightly higher accuracy, precision, recall, and F1-score compared to GPT-4o mini and LLaMA-4-Mav., the performance gap remains small. 
For computation-constrained AR deployments, GPT-4o mini and LLaMA-4-Mav. provide sufficient detection performance at substantially lower computation overhead and token budget.

\noindent\textbf{Privacy preservation effectiveness.} 
Table~\ref{part3_result} reports the privacy preservation effectiveness for PrivAR and alternative methods.
PrivAR achieves the highest CER of 95.71\%, indicating that its obfuscation method is most effective at preventing sensitive text from being accurately extracted by OCR. The PLR of PrivAR is 17.58\%, which is much lower than the PrivAR method without obfuscation (82.18\%). 
The oracle-guided obfuscation method achieves a slightly lower CER than PrivAR because it only obfuscates the annotated sensitive text, allowing OCR to accurately recognize the remaining content. In contrast, PrivAR obfuscates a broader region, 
which leads to a higher CER. Overall, PrivAR offers a practical balance between privacy and information retention.

\begin{table}[]
\centering
\caption{Privacy protection effectiveness of 3 methods.}
\label{part3_result}
\begin{tabular}{@{}ccc@{}}
\toprule
{Protection method} & {CER (\%)$\uparrow$}    & {PLR (\%)$\downarrow$} \\ \midrule
Oracle-guided obfuscation      & 91.11±14.57         & \textbf{2.29} \\ 
PriAR w/o obfuscation                 & 74.30±26.70         & 82.18            \\
PrivAR (ours)                       & \textbf{95.71±6.88} & 17.58            \\ \bottomrule
\vspace{-1cm}
\end{tabular}
\end{table}

\noindent\textbf{User study.} Participants' privacy concerns regarding AR technology were assessed before and after using PrivAR with a 7-point Likert scale (1 = strongly concerned, 7 = not concerned). Average scores decreased from 4.8 to 1.2, indicating substantially heightened privacy risk awareness following the study. Most participants (80\%) “strongly agree” or “agree” that PrivAR can promptly alert them to privacy risks. Among the three warning modes, both the region overlay warning and the center-screen warning were well liked, with each preferred by 40\% of participants. The region overlay warning was appreciated for identifying specific risk areas (but its usefulness depends on accurate alignment), while the center-screen warning was favored for its high visibility.

\section{Conclusion}
\label{sec:majhead}
PrivAR effectively addresses AR privacy challenges by obfuscating textual content at the edge server while preserving contextual cues for cloud-based VLM inference. This approach achieves high privacy risk detection accuracy (81.48\%) while 
reducing privacy leakage (17.58\% PLR). Our results establish a practical path for privacy-preserving VLM-based inference for AR systems.

\bibliographystyle{IEEEbib}

\bibliography{main}

\end{document}